\documentclass[conference]{IEEEtran}
\IEEEoverridecommandlockouts
\usepackage{amsthm}
\usepackage{amsmath}
\usepackage{amssymb}
\usepackage{booktabs}
\usepackage{cite}
\usepackage{graphics, framed}
\usepackage{graphicx}
\usepackage{epsfig}
\usepackage{epstopdf}
\usepackage{subfigure}
\usepackage{url}
\usepackage{multimedia}
\usepackage{paralist}
\usepackage[printonlyused]{acronym}
\usepackage{units}
\usepackage{balance}
\usepackage{multirow}
\usepackage{color,soul}
\usepackage[ruled,vlined]{algorithm2e}

\newcommand{\FGR}[1]{Fig.~\ref{#1}}

\newcommand{\SEC}[1]{Section~\ref{#1}}
\newcommand{\TAB}[1]{Table~\ref{#1}}
\newcommand{\EQ}[1]{(\ref{#1})}

\renewcommand{\baselinestretch}{0.88}
\acrodef{5G}[5G]{5\textsuperscript{th}-Generation}
\acrodef{BW}[BW]{bandwidth}
\acrodef{BS}[BS]{base station}
\acrodef{CW}[CW]{continuous wave}
\acrodef{D2D}[D2D]{device-to-device}
\acrodef{dB}[dB]{decibel}
\acrodef{dBi}[dBi]{decibel isotropic}
\acrodef{dBm}[dBm]{decibel over a milliwatt}
\acrodef{GAL}[GAL]{graph attention layer}
\acrodef{GAT}[GAT]{graph attention network}
\acrodef{Gbps}[Gbps]{gigabit per second}
\acrodef{GHz}[GHz]{gigahertz}
\acrodef{THz}[THz]{Terahertz}
\acrodef{RIS}[RIS]{reconfigurable intelligent surface}
\acrodef{CN}[CN]{core network}
\acrodef{PSK}[PSK]{phase shift keying}
\acrodef{QAM}[QAM]{quadrature amplitude modulation}
\acrodef{AWGN}[AWGN]{additive white Gaussian noise}
\acrodef{SNR}[SNR]{signal-to-noise ratio}
\acrodef{AF}[AF]{amplitude-and-forward}
\acrodef{MIMO}[MIMO]{multiple-input multiple-output}
\acrodef{mMIMO}[mMIMO]{massive-multiple-input multiple-output}
\acrodef{SDN}[SDN]{Software-defined network}
\acrodef{SON}[SON]{self-organizing network}
\acrodef{hetnet}[HetNet]{heterogeneous network}
\acrodef{FSO}[FSO]{free-space optics}
\acrodef{UM-MIMO}[UM-MIMO]{ultra-massive-MIMO}
\acrodef{AP}[AP]{access point}
\acrodef{UE}[UE]{user equipment}
\acrodef{NFP}[NFP]{networked flying platform}
\acrodef{UAV}[UAV]{unmanned aerial vehicle}
\acrodef{HAPS}[HAPS]{high-altitude platform station}
\acrodef{LEO}[LEO]{low-earth orbit}
\acrodef{BAN}[BAN]{body area network}
\acrodef{WLAN}[WLAN]{wireless local area network}
\acrodef{QoS}[QoS]{quality of service}
\acrodef{TCS}[TCS]{thermal control system}
\acrodef{QCL}[QCL]{quantum cascade laser}
\acrodef{CMOS}[CMOS]{complementary metal-oxide semiconductor}
\acrodef{V-HetNet}[V-HetNet]{vertical heterogeneous network}
\acrodef{DL}[DL]{deep learning}
\acrodef{DRL}[DRL]{deep reinforcement learning}
\acrodef{FDTD}[FDTD]{Finite-difference time-domain}
\acrodef{FEM}[FEM]{finite element method}
\acrodef{MoM}[MoM]{method of moments}
\acrodef{VNA}[VNA]{vector network analyzer}
\acrodef{CS}[CS]{channel sounder}
\acrodef{CIR}[CIR]{channel impulse response}
\acrodef{CTF}[CTF]{channel transfer function}
\acrodef{PN}[PN]{pseudo-noise}
\acrodef{TOA}[TOA]{time of arrival}
\acrodef{GMM}[GMM]{Gaussian mixture model}
\acrodef{OOK}[OOK]{on-off keying}
\acrodef{MLE}[MLE]{maximum likelihood estimation}
\acrodef{LOS}[LOS]{line-of-sight}
\acrodef{NLOS}[NLOS]{non-line-of-sight}
\acrodef{SG}[SG]{signal generator}
\acrodef{SA}[SA]{spectrum analyzer}
\acrodef{FDSOI}[FDSOI]{fully depleted silicon on insulator}
\acrodef{OpEx}[OpEx]{operational expenditures}
\acrodef{TDD}[TDD]{time division duplex}
\acrodef{CSI}[CSI]{channel state information}
\acrodef{MAC}[MAC]{medium access control}
\acrodef{GEO}[GEO]{geostationary orbit}
\acrodef{SWaP}[SWaP]{size, weight, and power}
\acrodef{NMSE}[NMSE]{normalized mean square error}
\acrodef{MSE}[MSE]{mean square error}
\acrodef{LS}[LS]{least square}
\ifCLASSINFOpdf
\else
\fi

\begin{document}

\title{Federated Learning for UAV-Based Spectrum Sensing: Enhancing Accuracy Through SNR-Weighted Model Aggregation}

\IEEEoverridecommandlockouts 


\author{\IEEEauthorblockN{K{\"{u}}r{\c{s}}at~Tekb{\i}y{\i}k\IEEEauthorrefmark{1},  G{\"{u}}ne{\c{s}}~Karabulut~Kurt\IEEEauthorrefmark{1}, Antoine~Lesage-Landry\IEEEauthorrefmark{2}}
\IEEEauthorblockA{\IEEEauthorrefmark{1}Department of Electrical Engineering, Polytechnique Montr\'eal, Poly-Grames Research Center, Montr\'eal, Canada\\ \IEEEauthorrefmark{2}Department of Electrical Engineering, Polytechnique Montr\'eal, GERAD \& Mila, Montr\'eal, Canada\\ Emails: \texttt{ktekbiyik@gmail.com, \{gunes.kurt, antoine.lesage-landry\}@polymtl.ca}}}

\maketitle

\begin{abstract}
The increasing demand for data usage in wireless communications requires using wider bands in the spectrum, especially for backhaul links. Yet, allocations in the spectrum for non-communication systems inhibit merging bands to achieve wider bandwidth. To overcome this issue, spectrum-sharing or opportunistic spectrum utilization by secondary users stands out as a promising solution. However, both approaches must minimize interference to primary users. Therefore, spectrum sensing becomes vital for such opportunistic usage, ensuring the proper operation of the primary users. Although this problem has been investigated for 2D networks, unmanned aerial vehicle (UAV) networks need different points of view concerning 3D space, its challenges, and opportunities. For this purpose, we propose a federated learning (FL)-based method for spectrum sensing in UAV networks to account for their distributed nature and limited computational capacity. FL enables local training without sharing raw data while guaranteeing the privacy of local users,lowering communication overhead, and increasing data diversity. Furthermore, we develop a federated aggregation method, namely \texttt{FedSNR}, that considers the signal-to-noise ratio observed by UAVs to acquire a global model. The numerical results show that the proposed architecture and the aggregation method outperform traditional methods.
\end{abstract}
\IEEEpeerreviewmaketitle
\acresetall

\section{Introduction}

As the demand for wireless communications increases, there is an indispensable need for more efficient radio spectrum utilization approaches. Considering the growing number of devices and the limited spectrum resources, spectrum-sharing has been proposed as a promising solution to meet the increasing demand by enabling both primary and secondary users to utilize the limited spectrum in a more efficient way. This would allow secondary users to access the spectrum without interfering with the primary users who are licensed to use the spectrum. For this purpose, various spectrum-sharing methods such as dynamic spectrum access and cognitive radio network~\cite{wang2010advances} have been recently investigated. As detailed in~\cite{wang2010advances}, the main challenge is to opportunistically access the spectrum while ensuring minimum interference with the primary users~\cite{zheleva2023radio, shang20203d}, as such users typically have licenses for the usage of dedicated spectrum bands. The interference observed in this band can significantly reduce the quality of their operation whether it is communication or active/passive remote sensing. Therefore, secondary users are obliged to use the spectrum with minimum interference to primary users. To this end, several studies have investigated spectrum sensing and accessing schemes to address this issue. Techniques such as energy detection, matched filtering, and feature detection have been studied in detail, but there are still challenges in achieving reliable spectrum sensing under different conditions~\cite{zhou2024hierarchical, choi2024spectrum}. Most of them have been proposed for the two-dimensional (2D) space without considering the opportunity coming from the three-dimensional (3D) space for multi-input multi-output (MIMO) systems or distributed networks such as unmanned aerial vehicle (UAV) swarms as illustrated in~\FGR{fig:system_model}. A novel aspect of this work is considering a 3D system for spectrum sensing, which is different from traditional 2D approaches that only consider time and frequency. Because a 3D system provides a more comprehensive understanding of the spectrum environment by incorporating spatial information (height or altitude), it enables better spectrum-sharing decisions in UAV communications. To address the distributed nature of spectrum sensing and sharing in UAV networks~\cite{lee2024federated}, federated learning (FL) is utilized in this study as it allows decentralized training of models across multiple devices or users without requiring them to share raw data. As mentioned in~\cite{shang2020spectrum}, FL is advantageous in spectrum-sharing scenarios where privacy concerns and distributed locations of users make centralized learning impractical. Furthermore, this study aims to consider local knowledge from multiple edge users while minimizing the need for computational resources at edge devices. As seen later, the communication overhead and the computational complexity at edge devices exhibit trade-offs and the capacity in terms of both need to be accounted for. The main contributions of this study are as follows: 

\begin{figure}[!t]
    \centering
    \includegraphics[width=\columnwidth, page=2]{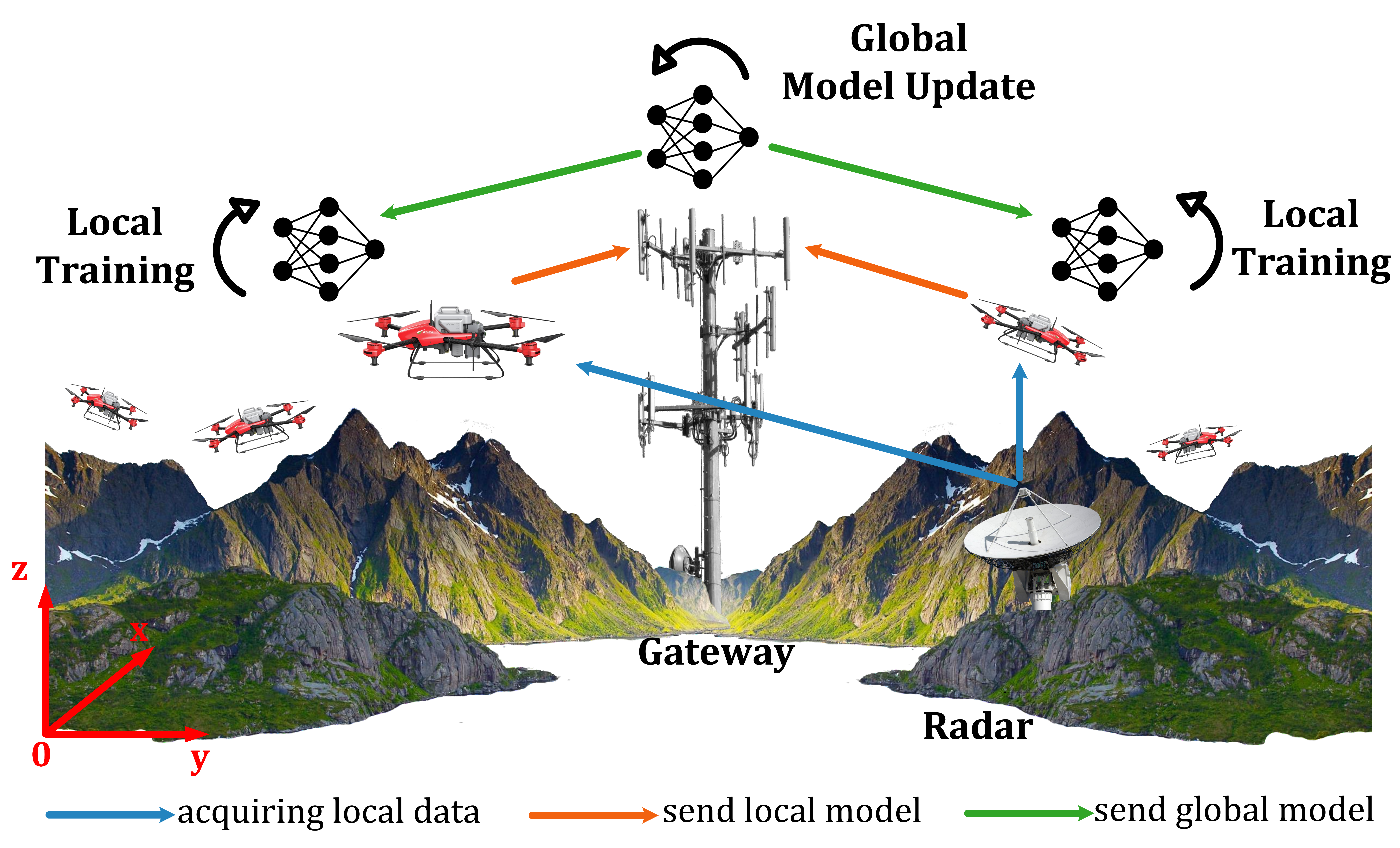}
    \caption{Distributed UAVs network employing FL for spectrum sensing.}
    \label{fig:system_model}
\end{figure}

\begin{enumerate}[{C}1]
    \item We present a lightweight convolutional neural network (CNN) for spectrum sensing tasks, optimized to process real and imaginary components of received signals in 3D environments incorporating spatial dimensions. 
    \item We employ FL and propose the \texttt{FedSNR} algorithm to effectively handle distributed learning for spectrum sensing. Numerical results show that our approach improves spectrum sensing accuracy. 
\end{enumerate}

The remainder of the paper is organized as follows: \SEC{sec:system_model} introduces the multi-UAV networks in a 3D environment. \SEC{sec:dl_model} describes the FL architecture and methodology, including the client model and federated aggregation methods. \SEC{sec:results} details the experimental setup and discusses the results. Finally, \SEC{sec:conclusion} concludes the paper with a discussion of the implications of this work and future research directions.


\section{System Model}\label{sec:system_model}

In this study, we consider the system model illustrated in~\FGR{fig:system_model} in which multiple UAVs operate within a 3D space. For convenience, we represent the space as a cuboid. Each UAV whose backhaul connection to the gateway utilizes spectrum opportunistically because of the high communication load is positioned in this space within the limits of its operational area. As a primary user of the spectrum, a radar is positioned at a fixed altitude and may use different types of radar waveforms. In this study, we consider five of them: continuous wave, frequency-modulated continuous wave, pulse, chirp, and phase-coded. For $i^{\text{th}}$ UAV, the received signal at time~$t$ is represented as follows:
\begin{equation}
    y_{i}(t) = \sqrt{\frac{P_\text{tx}}{PL_{i}\left(d_{i}^{t}, \theta_{i}^{t}\right)}} h_{i}(t)s(t)\mathrm{e}^{j2\pi f_\text{d}^{i} t} + n(t),
\end{equation}
where $P_\text{tx}$ is the transmit power of the legitimate user (i.e., radar). $PL_{i}$ is the path loss between $i^{\text{th}}$ UAV and radar, which is defined in \cite{al2017modeling} with considering distance and elevation angle jointly. The path loss is formulated in~\EQ{eq:path_loss}, where $d_{i}^{t}$ and $\theta_{i}^{t}$ are the distance and elevation angle between radar and $i^{\text{th}}$ UAV, respectively; $\alpha$, $\beta$, $\theta_0$, and $\zeta$ represent terrestrial path loss exponent, excess path loss scaler, angle offset, and angle scaler, respectively. The angles of transmission, referred to as elevation angles, are calculated based on the relative positions of the UAVs and the radar. These elevation angles and the distance between the radar and the UAVs, are critical for determining the path loss, which directly impacts the received signal power at the edge clients. 

The locations of the UAVs are assumed to follow a Matern Hardcore Point Process distribution~\cite{zhu2018secrecy} within the region bounded by a rectangular prism with dimensions ($x_\mathrm{max}$, $y_\mathrm{max}$, $z_\mathrm{max}$), representing the limits on the $x$-axis, $y$-axis, and $z$-axis, respectively. This distribution accounts for the strict minimum distance, $d_\mathrm{min}$, between UAVs due to safety regulations. As it ensures a minimum separation between UAVs, it reflects a realistic deployment scenario where UAVs shall maintain a sufficient distance to avoid interference. The radar's location is assumed to be uniformly distributed on a 2D plane at a fixed altitude, with its position bounded by the limits $x_\mathrm{max}$ and $y_\mathrm{max}$ on the $x$- and $y$-axes, respectively. Let the radar altitude be expressed as $z_\text{r}$. Thus, the distance and elevation angle between the radar and $i^{\text{th}}$ UAV are given as:
\begin{align}
    d_{i}^{t} &= \sqrt{(x_{i}^{t} - x_\text{r})^2 + (y_{i}^{t} - y_\text{r})^2 + (z_{i}^{t} - z_\text{r})^2} \label{eq:distance}\\
    \theta_{i}^{t} &= \arctan\left(\frac{z_{i}^{t} - z_\text{r}}{\sqrt{(x_{i}^{t} - x_\text{r})^2 + (y_{i}^{t} - y_\text{r})^2}}\right).
    \label{eq:elevation_angle}
\end{align}

Let $\nu_0$ be excess path loss offset, and $\mathcal{N}(0,\, \eta \theta_{i} + \sigma_0)$ be the normal distribution with zero mean and a standard deviation of $\eta \theta_{i} + \sigma_0$, where $\eta$ and $\sigma_0$ denote UAV shadowing slope and offset, respectively.
Our system model employs a small-scale fading channel, $h_{i}$, with a Rician distribution with a shape parameter of $K$.

\newcounter{mytempeqncnt20}
\begin{figure*}[ht!]
\normalsize
\setcounter{mytempeqncnt20}{\value{equation}}
\setcounter{equation}{3}
\begin{align}
       PL_{i}(d_{i}^{t}, \theta_{i}^{t}) = 10\alpha \log(d_{i}^{t}) + \beta (\theta_{i}^{t} - \theta_0)\mathrm{e}^{-\frac{\theta_{i}^t- \theta_0}{\zeta}} + \nu_0 + \mathcal{N}(0,\, \eta \theta_{i}^{t} + \sigma_0)
      \label{eq:path_loss}
\end{align}
\setcounter{equation}{\value{mytempeqncnt20}}
\hrulefill
\vspace*{1pt}
\end{figure*}
\setcounter{equation}{4}

Let $s(t)$ be the transmitted signal from the radar at time~$t$. Depending on its existence in the spectrum, the hypothesis for $n^{\text{th}}$ sample is defined as follows:

\begin{align}
\begin{split}
\mathcal{H}_0 &: \mathbf{s}[n] = 0 \\ 
\mathcal{H}_1 &: \mathbf{s}[n] = s\left(\frac{n}{f_\text{s}}\right),
\end{split}
\label{keq:hipotez}
\end{align}
where $f_\text{s}$ is the sampling frequency of the receiver's analog-to-digital converter. Finally, let $f_\text{d}^{i}$ be Doppler frequency characterized by the relative velocity, $v_{i}$, between transceivers as defined below:
\begin{align}
    f_\text{d}^{i}(t) = \frac{v_{i}(t)}{c}f_\text{c},
\end{align}
where $c$ and $f_\text{c}$ are the speed of light and carrier frequency, respectively. We assume that $v_{i}$ limited by $v_\mathrm{max}$ and uniformly distributed within $[0\,, v_\mathrm{max}]$.

\section{Federated Learning for Spectrum Sensing}\label{sec:dl_model}

In this section, we describe our FL approach and proposed aggregation method along with an edge model architecture. First, federated aggregation is introduced. Then, the edge model is explained in detail and the relation between architecture and the task is discussed. The last part of this section focuses to data generation and training processes. 

\subsection{Federated Aggregation}\label{sed:federated_learning}

As a decentralized machine learning technique, FL allows multiple edge devices to collaboratively train a global model with the coordination of a central server without sharing local data. This approach enables scalability, data privacy, and communication efficiency. Traditional methods for spectrum sensing with UAV swarm networks require gathering data from all devices in a central server, and raise concerns related to data privacy, latency, and communication overhead. Federated learning alleviates these challenges by training local models on each end-device, sharing only the model updates (i.e., gradients or weights) with the central server, which then combines them to form a global model~\cite{gafni2022federated}. At the core of this process is the aggregation method, which is crucial for obtaining an optimal global model. \texttt{FedAvg}~\cite{mcmahan2017communication}, a common aggregation method, averages the model updates (i.e., weights or gradients) from all participating devices to aggregate them into global model. In this approach, the central server does not receive raw data but instead accesses model parameters from each edge client and uses them to update the global model. Let $N$ be the number of UAVs, and $\mathbf{w}_{i}^{k}$ be the edge model parameters of the $i^{\text{th}}$ device at round $k$. The global model parameter $\mathbf{w}^{k+1}$ is updated on the central server by averaging the local models, weighted by the number of data samples on each device, as follows:
\begin{align}
    \mathbf{w}^{k+1} =  \frac{\sum_{i = 1}^{N} n_{i} \mathbf{w}_{i}^{k}}{\sum_{i = 1}^{N} n_{i}},
     \label{eq:fedavg}
\end{align}
where $n_{i}$ is the number of data samples on the $i^{\text{th}}$ UAV. After the aggregation step, the central server sends the updated global model $\mathbf{w}^{k+1}$ back to the edge devices for the next training round.

Besides \texttt{FedAvg}, we propose the \texttt{FedSNR} aggregation method by extending \texttt{FedAvg} and incorporating signal-to-noise ratio (SNR) as weights in the process. This is particularly important in UAV-based spectrum sensing, where signal quality varies significantly across devices. In the proposed approach, \texttt{FedSNR} prioritizes local parameters received from UAVs with higher SNR because it can be expected that these devices are likely to have more reliable data and less noise in their gradient updates. Let $\gamma_{i}$ be the SNR of the $i^{\text{th}}$ device. The \texttt{FedSNR} aggregation method substitutes the number of samples 
by the SNR, yielding
\begin{align}
    \mathbf{w}^{k+1} =  \frac{\sum_{i = 1}^{N} \gamma_{i} \mathbf{w}_{i}^{k}}{\sum_{i = 1}^{N} \gamma_{i}}.
    \label{eq:fedsnr}
\end{align}
The average SNR during sensing interval, $\gamma_{i}$, is defined as:
\begin{align}
   \gamma_{i} =  \frac{P_{i}}{N_{0\text{i}}}, 
   \label{eq:snr}
\end{align}
where $P_{i}$ is the average received power at the $i^{\text{th}}$ UAV during the sensing interval, $\frac{M}{f_\text{s}}$, where $M$ is the number of received signal samples used for spectrum sensing. Additionally, $N_{0i}$ is the noise floor, which might change from device to device due to the chipset designs. By employing SNR in the aggregation process, \texttt{FedSNR} which the global model better integrates information from noisy updates, which is crucial in dynamic environments like UAV networks where channel conditions change over time and space.

\subsection{Edge Model Architecture}\label{sec:edge_model}

\begin{figure*}[!t]
    \centering
    \includegraphics[width=\linewidth, page=3]{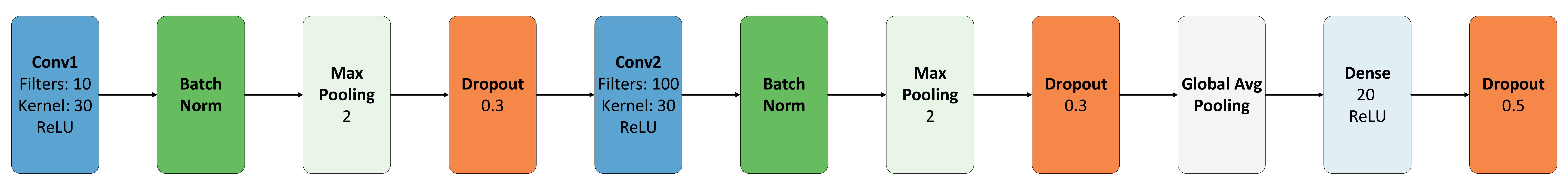}
    \caption{Illustration of edge model architecture consisting of two convolutional layers and several regularizations.}
    \label{fig:edge_model}
\end{figure*}

The edge model used in our federated learning framework is specifically designed to run efficiently on resource-constrained devices such as UAVs. Considering the limitations in computational capacity and memory of these devices, we aim to design a lightweight CNN architecture for processing time series data. To capture sequential patterns in the received signal, which is particularly useful for spectrum sensing tasks~\cite{tekbiyik2020robust}, we employ one-dimensional convolutions by using in-phase and quadrature parts of the signal as different channels. 

For the CNN's input layer, we use the received signal as input and represent it in terms of its real and imaginary components, as given:
\begin{align}
\mathbf{X} =
\begin{bmatrix}
\mathrm{Re(\mathbf{y}}[n]\mathrm{)} & \mathrm{Re(\mathbf{y}}[n-1]\mathrm{)} & \dots & \mathrm{Re(\mathbf{y}}[n-M+1\mathrm{])} \\
\mathrm{Im(\mathbf{y}}[n]\mathrm{)} & \mathrm{Im(\mathbf{y}}[n-1]\mathrm{)} & \dots & \mathrm{Im(\mathbf{y}}[n-M+1]\mathrm{)}
\end{bmatrix},
\end{align}
because it allows for real-valued operations that are computationally efficient and enable independent feature extraction from each component. $\mathbf{y}[n])$ is the $n^\text{th}$ sample of received signal during sensing period of $M$ samples.

The architecture includes first an initial convolutional layer consisting of $10$ filters with a kernel size of $30$, and ReLU activation function. The input data is processed by applying filters over temporal windows to allow the model to extract significant features from the data. In order to stabilize and speed up the training process, we apply batch normalization after the convolution layer. This layer improves convergence and model robustness by normalizing the output of the previous layer. Following batch normalization, we downsample its output while preserving the most important features by employing a max-pooling layer with a pool size of $2$ to reduce the dimensionality. Then, a dropout layer with a ratio of $0.3$ is introduced to avoid overfitting, especially critical given the limited data available at the edge. The second convolution layer again uses $100$ filters with a kernel size of $30$ and maintains the same padding to preserve the spatial dimensions of the input. This layer is important for learning more complex, higher-level features. As in the first convolution layer, batch normalization is applied to ensure stable learning, and then max-pooling operation is applied to further reduce the size of the data. The same dropout layer is used to maintain regularization throughout the learning process.

Instead of smoothing the output from convolutional layers, we adopt a global average pooling layer. This approach averages the feature maps along the temporal dimension, resulting in a less representative input sequence. As it significantly decreases the number of parameters in the model compared to traditional flattening operations, global average pooling might be preferred for edge devices to reduce the computational complexity and memory usage. Moreover, such pooling enhances the model generalization by focusing on the most salient features. The global pooling process is followed by a fully connected layer with $20$ neurons with the ReLU activation function. This dense layer further improves the model’s understanding of the input data by introducing nonlinear combinations of the extracted features. A dropout ratio of $0.5$ is applied to this layer to prevent overfitting. Overall, the edge model is illustrated in~\FGR{fig:edge_model}.

In sum, the model architecture is quite suitable for deployment on edge devices owing to the fact that it provides a balance between complexity and performance. By utilizing one-dimensional convolutions, a batch normalization, and a global average pooling, the model can efficiently process received signal samples with minimum computational and memory demands imposed on UAVs. Dropout layers further enhance the generalization ability of the model and help it perform well across varying data distributions encountered by different edge devices. It also minimizes the weight-diverging problem for each client that observes different wireless channels. This design allows each edge device to train locally on its own data, contributing to the FL process without excessive resource consumption.

\subsection{Dataset Generation and Training}\label{sec:training}

\begin{algorithm}[!t]
\SetAlgoLined
\SetKw{KwInitialization}{Initialization:}
\KwInitialization{Number of settings, $B$, $N$, $d_\mathrm{min}$, $x_\mathrm{max}$, $y_\mathrm{max}$, $z_\mathrm{max}$, $z_\mathrm{r}$, $P_\mathrm{tx}$, $N_0$, $f_\mathrm{c}$, $f_\mathrm{s}$, $M$, $v_\mathrm{max}$.}\\
\KwResult{Global model after federated aggregation.}

\For{$k \gets 1$ \textbf{to} number of settings} {
    $(\mathbf{x}, \mathbf{y}, \mathbf{z}) \gets$ 3D Matern Hardcore Process by using ($N, d_\mathrm{min}, x_\mathrm{max}$, $y_\mathrm{max}$, $z_\mathrm{max}$)\;
   $(x_\text{r}, y_\text{r}, z_\text{r}) \gets$  2D Uniform Distribution by using $(x_\mathrm{max}$, $y_\mathrm{max}$, $z_\text{r})$\;
    $\mathbf{d} \gets$ \EQ{eq:distance}\;
    $\mathbf{\theta} \gets$ \EQ{eq:elevation_angle}\;
    \For{$i \gets 1$ \textbf{to} $N$} {
        $ PL_{i}(d_{i}^{n}, \theta_{i}^{n}) \gets$ \EQ{eq:path_loss}\;
        $P_{i} \gets$ $\frac{P_\text{tx}}{P_{i}}$\;
        $\gamma_{i} \gets$ \EQ{eq:snr}\;
        
        \textbf{Generate signal data and labels}\;
        $(\mathbf{X}, \mathbf{y}) \gets$ Training Batch for $i^{\text{th}}$ UAV\;
        
        \textbf{Create and train individual edge models}\;
        $edge\_model \gets$ Build Edge Model\;
        $\mathbf{w}_{i}^{k} \gets$ Train Edge Model\;
        $\mathbf{w}^{k} \gets$ append $\mathbf{w}_{i}^{k}$\;
    }
    \textbf{Aggregate edge models into a global model}\;
    $\mathbf{w}^{k} \gets$ FA given in \EQ{eq:fedavg} or \EQ{eq:fedsnr}\;
}
\Return{$global\_model, \mathbf{w}^{k}$}
\caption{FL with UAV Signal Dataset Generation}
\label{algo:fl_data}
\end{algorithm}

As described above, we propose a new approach to spectrum sensing in UAV networks using FL and leverage a dataset specifically created for this task. The dataset is created by simulating a realistic 3D environment for UAVs and radar systems and then training individual models at the end nodes, which are then aggregated into a global model via FL. 

The generation process begins with defining the simulation parameters; the number of settings, number of data per UAV ($B$), number of UAVs ($N$),  simulation boundaries in $x$, $y$, and $z$ dimensions, minimum distance between UAVs, radar altitude, transmit ($P_\text{tx}$) and noise power ($N_0$), signal length ($M$), sampling frequency ($f_\text{s}$), speed range ($v_\mathrm{max}$), and carrier frequency ($f_\text{c}$). 

In each setting, a 3D point cloud representing UAV locations is created using a Matern Hardcore Process that guarantees spatial randomness while maintaining minimal separation between UAVs. The radar is positioned at a fixed altitude and its coordinates are uniformly distributed on a 2D plane. The distances and elevation angles between the UAVs and the radar are calculated based on their relative positions. These parameters are then used to calculate the path loss, which determines the received signal strength for each UAV.

After the path loss is obtained, the received signal strength is calculated and the SNR in dB scale, is derived by subtracting the noise power. The signal data for each UAV is generated using these parameters with signal samples generated by employing the different radar waveforms: continuous wave, frequency-modulated continuous wave, pulse, chirp, and phase-coded. The radar waveform is selected randomly. Each group consists of wireless channel- and velocity-related features (i.e., Doppler) sampled at the specified frequency and labeled accordingly.

Each UAV operates as an edge node in the FL system. Individual edge models are generated using CNNs as described above, whose input shape is determined by the signal length~$M$. These models are trained on locally generated groups using the prepared data. Each client trains the local models for 20 epochs with an early stopping criteria if the training process saturates and then optimizes the weights via Adam optimizer with respect to binary cross entropy loss. The updated weights are then sent to the central server for aggregation. 

After each training round, the trained weights from each edge model are combined into a global model. This aggregation process provides a better global model that benefits from decentralized training while ensuring data privacy and low overhead for each UAV node. Algorithm~\ref{algo:fl_data} summarizes the data generation, local training, and federated aggregation processes. The data generation and federated training method provides a scalable solution for spectrum sensing in UAV networks, especially in scenarios with large numbers of UAVs and complex environmental conditions.


\section{Results and Discussions}\label{sec:results}

\begin{table}[!t]
\centering
\caption{Simulation parameters for UAV-based spectrum sensing system. Default values are shown in \textbf{bold}.}
\begin{tabular}{c c c}
\toprule
Parameter & Notation & Value \\ \midrule
\# Settings    &    --     & $500$ \\
Data per UAV   &    B   & $\mathbf{256}$  \\
\# UAVs  & $N$                  & $\mathbf{16}$ \\
Rician Parameter  &         $K$               & $\mathbf{10}$ \\
Transmit Power  &         $P_\text{tx}$               & $\mathbf{5}$ dB \\
Noise Power  &         $N_0$            & $-93$ dBm \\
Signal Length  &         $M$            & $3000$ \\
Sampling Frequency  &         $f_\text{s}$            & $300$ MHz \\
Carrier Frequency  &         $f_\text{c}$            & $10$ GHz \\
$x$-axis Limit  &         $x_\mathrm{max}$            & $5000$ m \\
$y$-axis Limit  &         $y_\mathrm{max}$            & $5000$ m \\
$z$-axis Limit  &         $z_\mathrm{max}$            & $120$ m \\
Radar Altitude  &         $z_\mathrm{r}$            & $40$ m \\
Minimum Distance  &         $d_\mathrm{min}$            & $100$ m \\
Maximum UAV Speed  &         $v_\mathrm{max}$            & $44$ m/s \\
Path Loss Exponent  &         $\alpha$           & $3.04$ \\
Angle Offset  &         $\theta_0$           & $-3.61$ \\
Excess Path Loss Scaler  &         $\beta$           & $-23.29$ \\
Angle Scaler  &         $\zeta$           & $4.14$ \\
Excess Path Loss Offset  &         $\nu_0$           & $20.70$ \\
Shadowing Slope  &         $\eta$           & $-0.41$ \\
Shadowing Offset &         $\sigma_0$           & $5.86$ \\ \bottomrule
\end{tabular}
\label{tab:sim_params}
\end{table}

In this section, we present and discuss the numerical results obtained from the proposed UAV-based spectrum sensing system using FL. The results are discussed in terms of the overall performance of the system, and its accuracy under varying environmental and operational conditions.

The simulation considered 16 UAVs deployed in a 3D space bounded by $5000$ meters along the $x$- and $y$-axes and $120$ meters along the $z$-axis, while the radar is positioned at a fixed altitude of $40$ meters. Each UAV collects $256$ samples per setting, and $500$ different settings are simulated. The transmit power is set to $5$ dB, and the noise floor is assumed to be $-93$~dBm. The Rician fading parameter $K = 10$ and the carrier frequency of $10$ GHz are used to reflect real-world communication conditions in urban environments. We set the sensing duration $10\, \mu$s, i.e., $M = 3000$ for $f_\text{s} = 300$ MHz. For the parameters related to the path loss model, we adopt the default values given in~\cite{al2017modeling}. The experiments are conducted with the simulation parameters summarized in~\TAB{tab:sim_params}. 

The \texttt{FedAvg} algorithm is used as a baseline for comparison with the proposed \texttt{FedSNR}, which utilizes the SNR of each UAV in the aggregation process. This setting ensures that model updates from UAVs with better signal quality receive more weight during the global model update process. We train each edge model independently by using data concatenated from each setting. Then, we test the models and evaluate the average of the independent sensing performance of all UAVs for comparison between FL and the traditional approach, namely individually trained edge models, which are employed as the benchmark. 

First, we investigate the relation between the transmit power and the sensing accuracy. \FGR{fig:acc_vs_pt} shows the spectrum detection accuracy against the transmission power. \texttt{FedAvg} and \texttt{FedSNR} show improved accuracy when increasing transmission power because of the improved signal quality. However, the \texttt{FedSNR} outperforms \texttt{FedAvg} at all power levels. At low power settings, \texttt{FedSNR} achieves a significant $4\%$ increase in accuracy. These results illustrate its ability to process weaker signals by prioritizing updates from UAVs with a higher SNR. When the transmit power increases above $15$~dB, a sufficient SNR is achieved at all edge nodes; hence, reducing the advantage of \texttt{FedSNR} over \texttt{FedAvg}. For comparison purposes, we train each edge model without sharing the weights and then test them individually.  \FGR{fig:acc_vs_pt} also shows the average accuracy of local models, which is much lower than the FL methods.

\begin{figure}[!t]
    \centering
    \includegraphics[width=\columnwidth]{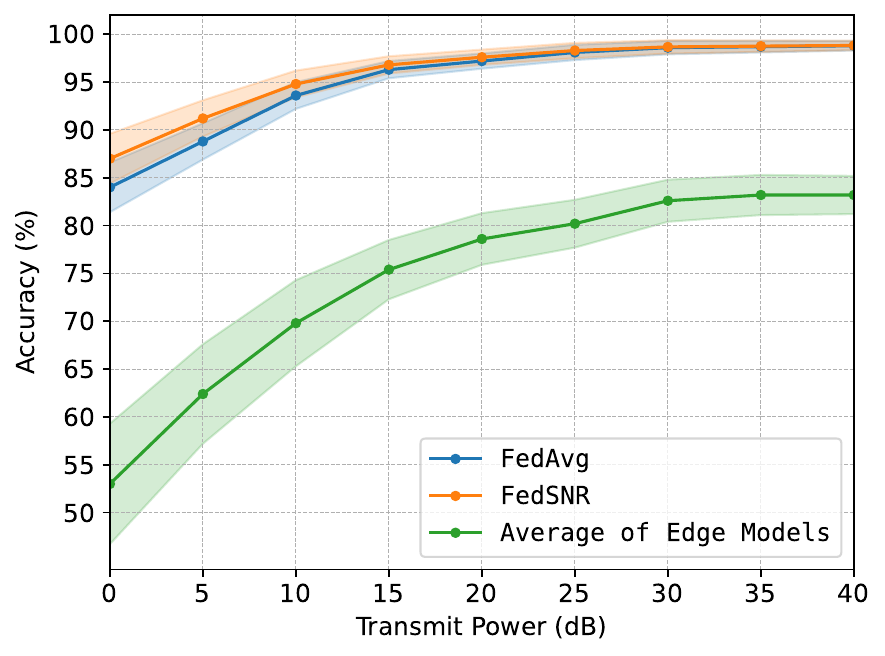}
    \caption{Sensing accuracy under different transmit power. The green plot shows the average accuracy of independently trained and tested edge models. The shadowed regions represent $95\%$ confidence intervals.}
    \label{fig:acc_vs_pt}
\end{figure}

Next, we analyze the impact of the UAV number on the accuracy in~\FGR{fig:acc_vs_num_uavs}. Because data from multiple UAVs provides a richer dataset from the spectrum, e.g., incorporating different wireless channel characteristics, increasing the number of UAVs generally improves the overall sensing accuracy. \FGR{fig:acc_vs_num_uavs} shows that \texttt{FedSNR} has a clear advantage over \texttt{FedAvg} in this example. When fewer UAVs are deployed, it is more likely that they end up positioned far away from the radar, resulting in a poor signal quality, and in turn reducing the learning performance because the raw data is becoming similar to noise. This setting implies that \texttt{FedSNR} is more effective than \texttt{FedAvg} in distributed systems with limited edge nodes. For dense deployments, \texttt{FedAvg}'s performance approaches \texttt{FedSNR}’s. For example, The accuracy of \texttt{FedSNR} stabilizes at a level close to $97.6\%$ for $16$ UAVs, while \texttt{FedAvg} achieves a slightly lower accuracy at $97.2\%$ under the same conditions.

\begin{figure}[!t]
    \centering
    \includegraphics[width=\columnwidth]{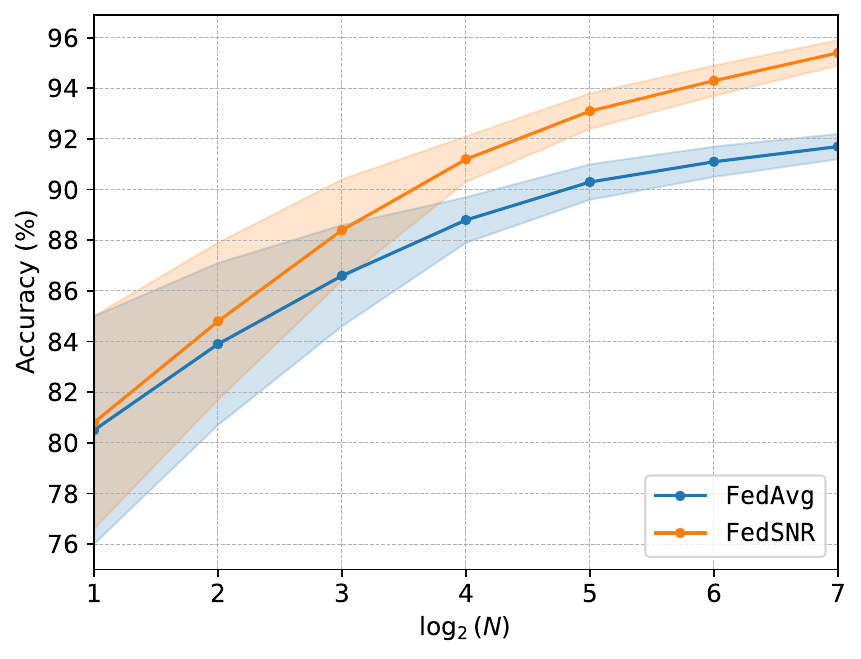}
    \caption{Spectrum sensing accuracy under varying number of UAVs.}
    \label{fig:acc_vs_num_uavs}
\end{figure}

As UAVs might experience line-of-sight (LoS) and none-line-of-sight (NLoS) channels due to higher mobility, the impact of channel conditions needs to be analyzed. As described above, the system model employs Rician fading channels with shape parameter, $K$, which denotes the power ratio of LoS and NLoS components. \FGR{fig:acc_vs_rician_params} shows the sensing performance under different conditions. As higher values of $K$ correspond to stronger LoS conditions and provide higher signal quality, sensing accuracy improves when LoS component dominates. Again, \texttt{FedSNR} shows better performance at lower values of $K$, where channel conditions are less stable due to significant NLoS components, compared to the other aggregation method. This result also illustrates the adaptability of the proposed method when overcoming varying channel conditions, which is a critical advantage in UAV networks where mobility and urban environments might create variability in signal quality. 

\begin{figure}[!t]
    \centering
    \includegraphics[width=\columnwidth]{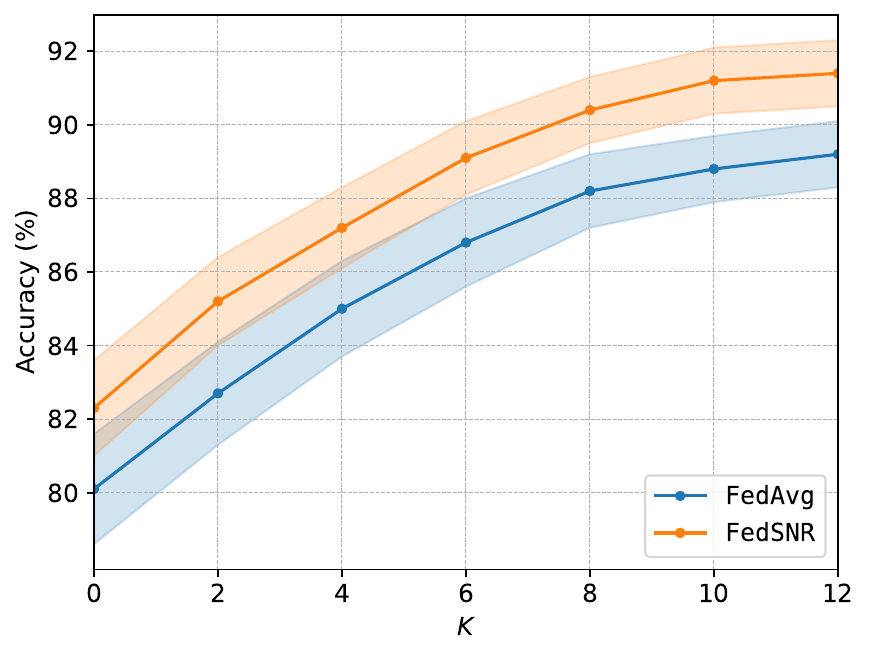}
    \caption{Impact of Rician fading parameter $K$ on spectrum sensing accuracy.}
    \label{fig:acc_vs_rician_params}
\end{figure}

Finally, we investigate the effect of the dataset size per UAV which is essential for analyzing the trade-off between computational complexity and accuracy for resource-constrained edge devices. To this aim, we analyze the sensing accuracy for varying dataset sizes per UAV in~\FGR{fig:acc_vs_batch_size}. As the available training data for each UAV increases in size, both \texttt{FedAvg} and \texttt{FedSNR} show improvements in accuracy, as more datas allows each UAV to calculate more reliable local model updates. Conversely, a decrease in the number of training data per UAV brings serious challenges. With smaller datasets, each UAV calculates less generalizable local model weights and leading to significant inconsistencies in the model weights then sent to the gateway for aggregation after each round. This variation in client weights results in the global model fluctuating from round to round, making it difficult to converge to optimal global weights. Fluctuations in model weights across UAVs prevent the aggregation method from stabilizing the global model and it leads to inconsistent learning.  performance across rounds. A smaller amount of training data per UAV increases the difficulty of finding an optimal global model. Large differences in local model updates hinder the aggregation process, prevent the system from reaching convergence, and thereby, reduce the overall effectiveness of the learning process. This is frequently experienced when edge nodes observe non-independent and identically distributed data. This phenomenon, known as weight divergence, requires dedicated approaches to deal with~\cite{ zhao2021federated}. This result highlights that UAVs should have access to sufficient training data to create stable and accurate model updates for the FL system.

\begin{figure}[!t]
    \centering
    \includegraphics[width=\columnwidth]{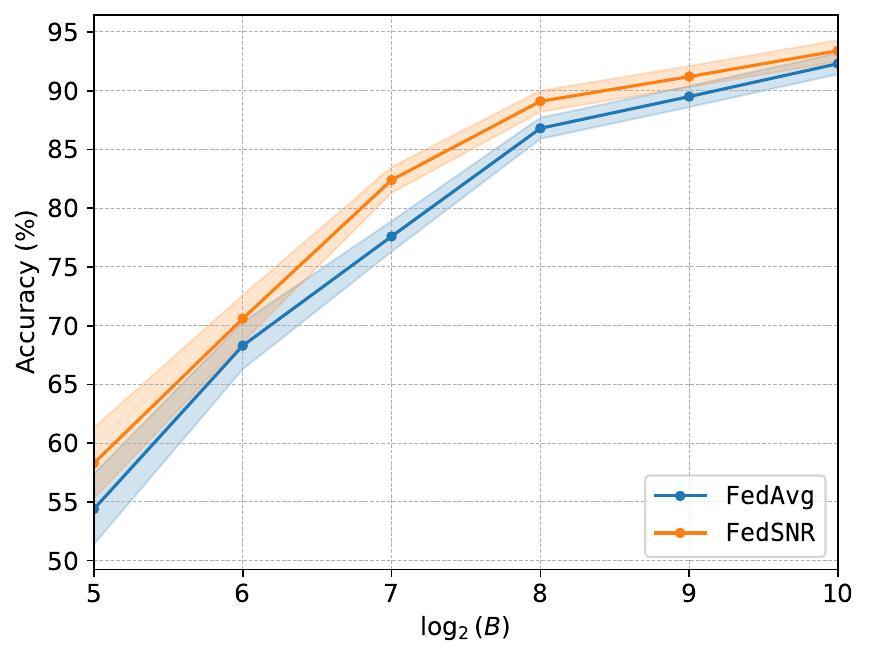}
    \caption{Spectrum sensing accuracy under varying dataset size per UAV.}
    \label{fig:acc_vs_batch_size}
\end{figure}

In sum, the scalability of the proposed system is another important factor. As shown in the numerical results because \texttt{FedSNR} performs well even with fewer UAVs and smaller batch sizes, it is efficient for resource-constrained networks. Its ability to maintain high accuracy with minimal computational and communication overhead makes it practical for real-time applications in both large-scale and small-scale deployments. Beyond spectrum sensing, the proposed system can be easily extended to other applications where decentralized learning and dynamic environments are key considerations such as vehicular and IoT networks.


\section{Concluding Remarks}\label{sec:conclusion}

This study proposes a spectrum-sensing system for UAV networks that opportunistically utilizes the spectrum band dedicated to radar systems to alleviate increasing demands in wireless communications. Considering the distributed nature of UAV networks, a federated approach is adopted. Numerical results show that the proposed FL approach can achieve higher accuracy than traditional local training. We propose a federated aggregation method, \texttt{FedSNR}, incorporating SNR observed by UAVs to avoid misleading weights from data with poor signal quality. The results show that the proposed aggregation method outperforms the widely used aggregation method, \texttt{FedAvg}. However, weight divergence remains an open issue that needs to be handled in a specialized way, considering the non-independent and non-identically distributed signals due to the nature of UAV networks experiencing different wireless channels and mobility. This is a topic for future works that could enable local training without weight divergence issues on memory-constrained edge devices. While it is not a limitation for UAVs, a new approach dedicated to small datasets would be beneficial, as our proposed architecture is also suitable for IoT networks that may face challenges due to memory limitations at the edge.
\balance
\bibliographystyle{IEEEtran}
\bibliography{main}
\end{document}